\pdfoutput=1

\documentclass[11pt]{article}

\usepackage{acl}

\usepackage{times}
\usepackage{latexsym}

\usepackage{multirow}
\usepackage{graphicx}
\usepackage{colortbl}
\usepackage[T1]{fontenc}

\usepackage[utf8]{inputenc}
\usepackage{svg}
\usepackage{microtype}

\usepackage{inconsolata}
\usepackage[export]{adjustbox}

\title{A linguistically-motivated evaluation methodology for unraveling model's abilities in reading comprehension tasks}

\author{Elie Antoine\textsuperscript{1}\ , Frédéric Béchet\textsuperscript{1, 4}\ , Géraldine Damnati\textsuperscript{2}\ , Philippe Langlais\textsuperscript{3}\\
\textsuperscript{1}CNRS, LIS, Aix-Marseille Université, France \; \footnotesize{\texttt{\{first.last\}@lis-lab.fr}}\\
\textsuperscript{2}Orange Innovation, DATA\&AI, Lannion, France \; \footnotesize{\texttt{\{first.last\}@orange.com}}\\
\textsuperscript{3}RALI, DIRO, Université de Montréal, Canada \; \footnotesize{felipe@iro.umontreal.ca}\\
\textsuperscript{4}International Laboratory on Learning Systems (ILLS - IRL CNRS), Montreal\\
}

\newcommand\felipe[1]{}
\newcommand\calor{\textsc{CALOR}}

\begin{document}
\maketitle

\begin{abstract}
We introduce an evaluation methodology for reading comprehension tasks  based on the intuition that certain examples, by the virtue of their linguistic complexity, consistently yield lower scores regardless of model size or architecture. We capitalize on semantic frame annotation for characterizing this complexity, and study seven complexity factors that may account for model's difficulty.  We first deploy this methodology  on a carefully annotated French reading comprehension benchmark showing that two of those complexity factors are indeed good predictors of models' failure, while others are less so. We further deploy our methodology on a well studied English benchmark by using ChatGPT as a proxy for semantic annotation. 
Our study reveals that fine-grained linguistically-motivated  automatic evaluation of a reading comprehension task is not only possible, but helps understand models' abilities to handle specific linguistic characteristics of input examples. It also shows that current state-of-the-art models fail with some for those characteristics  which suggests that adequately handling them requires more than merely increasing model size.
\end{abstract}

\section{Introduction}

Generative language models, and very large ones in particular, define the current state-of-the-art in a number of Natural Language Processing tasks.  Yet, despite the impressive quantity of scientific studies dedicated to them, the capabilities, limitations, and risks of these models remain largely unknown. 

In this work, we argue that black-box evaluations across various tasks, datasets, and languages~\cite{liang2023holistic,srivastava2022beyond} is not enough to portrait current models abilities and instead propose in Section~\ref{sec:method} a linguistically fine-grained evaluation methodology that  capitalizes on semantic frame annotation \cite{baker1998berkeley} to characterize examples thanks to a small number of  complexity factors  we describe in Section~\ref{sec:complexity-factors}.

Question Answering (QA) from documents has been extensively studied since the advent of deep neural network-based models, facilitated by large evaluation corpora such as SQuAD~\cite{D16-1264} and MultiRC~\cite{khashabi2018looking}, part of the SuperGLUE benchmark~\cite{wang2019superglue}. Transformer-based models consistently top leaderboards\footnote{\url{https://rajpurkar.github.io/SQuAD-explorer},\url{https://super.gluebenchmark.com/leaderboard}}, outperforming humans. For a nuanced view, see the position paper by \cite{tedeschi-etal-2023-whats}, but we acknowledge this belief as highlighting the challenge of evaluating QA, due to the subjective nature of answer generation and models capturing training data biases \cite{mccoy-etal-2019-right}. Thus, QA offers an interesting playground of our evaluation method that we consider here.

As a proof of concept, we apply our methodology to a publicly available reading comprehension benchmark \calor~\cite{bechet2019calor}, which includes French Question-Answer pairs with detailed semantic annotations on the relation linking questions and answers. We demonstrate that certain complexity factors can effectively predict model limitations, regardless of size or architecture. In Section~\ref{sec:natural_qa}, we extended our methodology to the NaturalQA~\cite{kwiatkowski2019natural} benchmark, using ChatGPT to compute complexity factors. Our results show that models of various sizes and architectures struggle with certain examples, suggesting that addressing these challenges requires more than just scaling up model size. By presenting a method to automatically select these challenging examples, we provide a means for monitoring further progress in reading comprehension.
The data used and collected in this study is available on the following link: \url{https://anonymous.4open.science/r/CALOR-QA-EMNLP}.

\section{Method}

\label{sec:method}

Our goal is twofold: first, to partition an evaluation corpus into several subsets, each with a distinct (linguistic) level of complexity; and second, to identify linguistically motivated factors that explain the variations in complexity across these subsets. We partition examples based on the analysis of systems' output inspired by the ROVER method~\cite{fiscus1997post}. To ensure independence from any single model when doing so, we propose using a set of models $M=m_1,m_2,\ldots,m_n$ adapted to perform the task and bin examples according to the number of models that agree in their answer with the majority vote. Thus, examples are partitioned  into $n$ bins (from total disagreement to full agreement); partition 1 grouping examples where all $n$ systems' outputs differ, while partition $n$ gather examples where all systems agree. 

To explain why some subsets are more complex than others, we confront linguistic assumptions formulated as complexity factors to examples  in each bin, proceeding as follows: 

\begin{enumerate}

\item We formulate several assumptions about semantic complexity factors ($F={f_1,f_2,\ldots}$) as binary questions applicable to examples in the evaluation corpus. For instance: \textit{Does finding the answer require solving a coreference chain?}

\item For each factor $f$, we divide the evaluation corpus into two subsets based on whether the examples answer ``yes'' ($E_{f}$=\textit{difficult subset}) or ``no'' ($\bar{E}_{f}$=\textit{easy subset}) to the question posed by the factor. When a binary factor requires a threshold to effectively divide the corpus (as in \textit{is the value corresponding to the factor higher (``yes'') than the threshold or not (``no'')?}) we use quantitative data to set this threshold in order to ensures a balanced division of the corpus.

\item For each factor $f$ and model $m$, we compute the performance of model $m$ on partitions $E_{f}$ and $\bar{E}_{f}$: $S(m,E_{f})$ and $S(m,\bar{E}_{f})$, and compute $\delta(m,f)$, a score which quantifies  performance degradation of model $m$ due to complexity factor $f$ as $\lfloor (S(m,E_{f})-S(m,\bar{E}_{f}))*100\rfloor$.

\item Finally, we calculate a measure of statistical significance for $\delta(m,f)$ with the Mann-Whitney $U$ test with a 5\% risk level between the two partitions $E_{f}$ and $\bar{E}_{f}$. This test takes into account the value of $\delta(m,f)$ and the characteristics of each set in the partition.

\end{enumerate}

As stated in the introduction, we applied our method to a reading comprehension task, which involves a QA process based on documents. The complexity factors we evaluate in this study were defined through a controlled experiment on the CALOR evaluation corpus, which was manually annotated with semantic frames and enriched with QA based on these frames. This process is described in the next section.

\section{Semantic complexity factors}
\label{sec:complexity-factors}

\subsection{A semantically-controlled QA corpus}
\label{sec:corpus}
We use the publicly available \calor{} corpus \cite{marzinotto:hal-01959187}
which contains documents semantically annotated with the Berkeley Framenet semantic model. This corpus includes French texts from Wikipedia as well as a collection of historical documents covering three main themes: First World War, archaeology, and antiquity.
The semantic annotation of this corpus consists of Semantic Frames that describe prototypical situations (e.g., decide, lose, attack, defeat). A trigger word of the Frame, called the Lexical Unit (LU), is identified, followed by the specification of the arguments, known as Frame Elements (FE). 

\begin{figure}[] 
\centering
\includegraphics[width=0.5\textwidth]{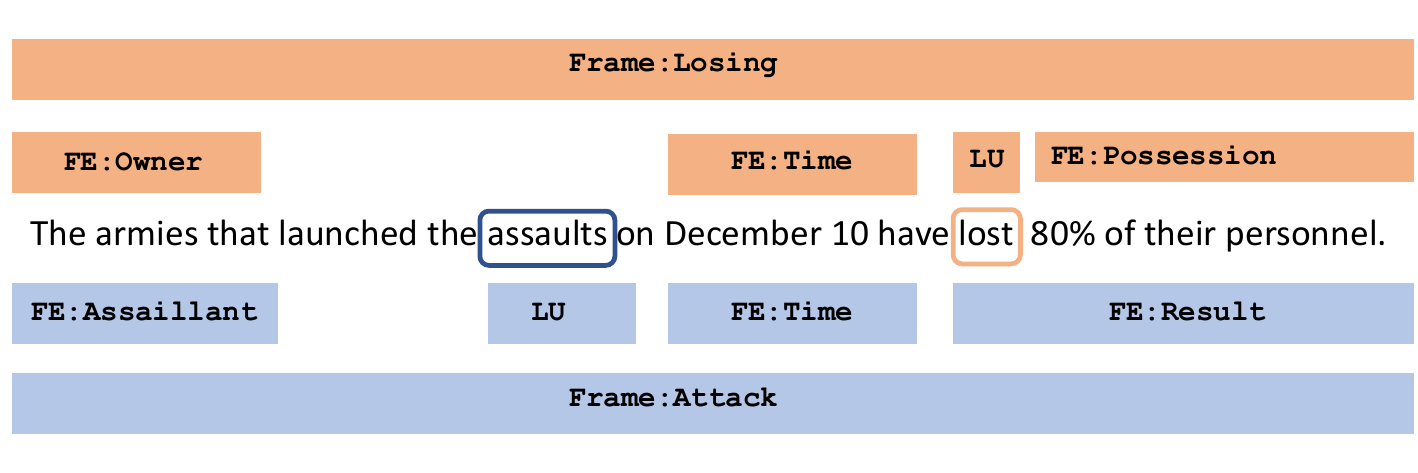}
\caption{Example of sentence annotated with two semantic frames}
\label{fig:framenet}
\end{figure}

In \cite{bechet2019calor}, it was enhanced with semantically controlled question-and-answer examples. This process involved selecting a semantic Frame and a corresponding FE from sentences, then having annotators generate questions whose answers were the selected Frame Elements, with the remaining elements providing context. By varying these selections, a dataset of questions, answers, and their semantic classes was created. Coreference chains were also annotated when needed. This approach produced a corpus of 1785 questions from 54 semantic frames, serving as a valuable resource for validating our methodology under controlled conditions.
An example of an annotated sentence from the corpus is shown in Figure \ref{fig:framenet}. Based on these two frame annotations, annotators could have formulated several questions, such as: \textit{"(1) Who lost the majority of their troops on December 10?"} or \textit{"(2) Who started the attack on December 10?"} In both instances, the sentence provides the answer "armies." However, the correct answer, derived from resolving the coreference chain in the paragraph, is "Central Empire coalition."

\subsection{Designing complexity factors}

\begin{figure*}[ht]
\includegraphics[width=1.5\columnwidth,center]{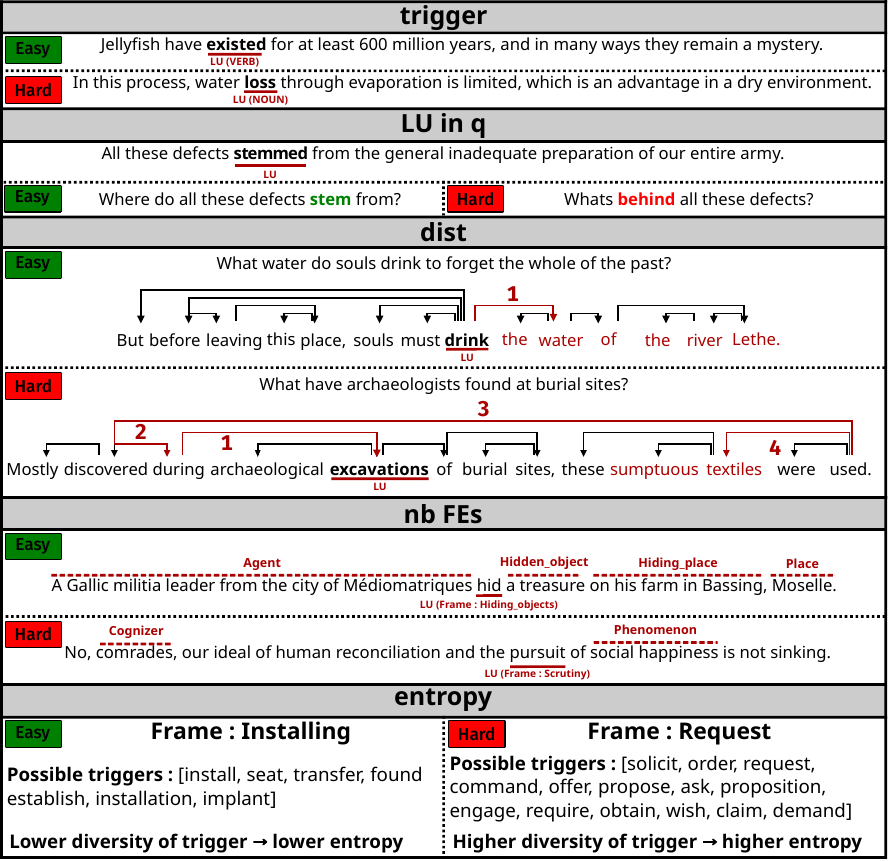}
\caption{Example of some complexity factors considered}
\label{fig:all_factors}
\end{figure*}

We consider in this study three types of factors based on the semantic frame annotation available in the corpus: factors capturing potential training biases ($f_{bias}$); factors based on lexical choices and syntactic structures of QA examples ($f_{coref}$, $f_{trigger}$, $f_{LU\;in\;q}$ $f_{dist}$) inspired by complexity factors proposed for automatic parsing of frames in~\cite{marzinotto:hal-01731385}; finally factors linked intrinsically to the semantic relation defined by a frame ($f_{nb\;FEs}$ and $f_{entropy}$). Here's a concise overview of these factors, with examples for all but $f_{bias}$ and $f_{coref}$ presented in Figure~\ref{fig:all_factors}.

\paragraph{$f_{bias}:$ bias in the training/adaptation corpus.} In the experiment section, we  use the French QA corpus FQuAD \cite{dhoffschmidt-etal-2020-fquad} for adapting several models to the QA task. This complexity factor explores the relationship between the frame distribution in this adaptation corpus and the model scores in the evaluation corpus.
To explore this factor, we used the tool described in \cite{marzinotto-etal-2019-robust} to automatically annotate the text data (context) of the FQuAD adaptation corpus with Frames and estimated the frequency of each Frame.
We then defined two sets of Frames: $F+$ for the more frequent Frames and $F-$ for the less frequent ones. The set $E_{f}$ consists of QA examples based on Frames in $F-$ (the rare ones), while $\bar{E}_{f}$ includes those based on Frames in $F+$ (the common ones).

\paragraph{$f_{coref}:$ coreference.} The need to resolve a coreference is a potential complexity factor. As mentioned in Section~\ref{sec:corpus}, coreference chains are annotated for the arguments of the semantic relations linking questions and answers, allowing us to divide the test corpus in two parts: examples with a coreference chain to be resolved to find the answer $E_{f}$ and the others $\bar{E}_{f}$. Both examples of question given for figure~\ref{fig:framenet} belongs to $E_{f}$ as a coreference resolution is needed to find the answer.

\paragraph{$f_{trigger}:$ nature of the semantic relation trigger.} The triggers of a frame in the FrameNet model, called \textit{Lexical Unit - LU}, can be either verbal or nominal.
It has been shown~\cite{marzinotto:hal-01731385} that relations triggered by a nominal LU are more difficult to process. We therefore divide the examples in the evaluation corpus according to the nature of the LU: either nominal $E_{f}$, or verbal $\bar{E}_{f}$.

\paragraph{$f_{LU\;in\;q}:$ presence of the frame trigger in the question.} When the same term triggers the semantic relationship in the context and in the question, the example is intuitively simpler to treat. To capture this, we bin examples in subset $E_{f}$ where the trigger is different between the question and context, and in $\bar{E}_{f}$ otherwise.

\paragraph{$f_{dist}:$ syntactic distance between the frame trigger and the answer.} The syntactic distance between the frame trigger and the answer may potentially challenge models as a greater distance may increase ambiguity for finding the answer to the question. We calculate the distance in terms of dependency arcs through a syntactic analysis of the corpus\footnote{We used the spaCy toolkit: \url{https://spacy.io}} and group together examples with at least  two dependency arcs between the trigger and the response in the subset $E_{f}$, and group those with only one arc in $\bar{E}_{f}$.

 \paragraph{$f_{nb\;FEs}:$ number of arguments in the frame.} Certain semantic relations exhibit varying numbers of Frame Elements (FEs). The number of FEs within the semantic relation underlying a question-answer pair can influence model efficiency: a higher number of FEs provides a richer contextual basis for accurately identifying the answer, while a smaller number of FEs can make the task more ambiguous. We categorize examples with no more than two annotated FEs into the subset $E_{f}$, and those with more than two FEs into $\bar{E}_{f}$. Our focus is on the manually annotated FEs present in the context of the question, rather than the theoretical number of FEs for the frame in Berkeley FrameNet.

\paragraph{$f_{entropy}:$ measure of entropy in the distribution of LUs for a given frame.} Some frames are consistently triggered by the same terms, while others exhibit much greater diversity, leading to ambiguity in their triggers. This measure of 'surprise' can be quantified through the entropy of the LU distribution in the evaluation corpus. A higher entropy suggests increased ambiguity in frame triggering. We include examples in the subset $E_{f}$ for frames with an entropy value above a threshold $\alpha$, and in $\bar{E}_{f}$ for frames below the same threshold, calculated as the median entropy value across all frames.

\section{Controlled experiment}
\label{sec:controlled_qa}

We compare seven pre-trained language models: one is a classification model based on a BERT architecture \cite{devlin-etal-2019-bert} developed for the French language, CamemBERT \cite{martin2020camembert}; three models are multilingual generative models based on \textbf{T5} (T5-LARGE, FLAN-T5-LARGE \cite{wei2021finetuned}, MT5-LARGE \cite{xue2020mt5}), and three models are current Large Language Models (LLMs): LLAMA2 \cite{touvron2023llama}, Mixtral 8x7B \cite{jiang2024mixtral} and ChatGPT-3.5\footnote{API from \url{https://chat.openai.com}}.

All these pre-trained models, except ChatGPT3.5 and Mixtral 8x7B, have been adapted to our QA task using the French corpus FQuAD \cite{dhoffschmidt-etal-2020-fquad}. This corpus, constructed similarly to SQuAD \cite{D16-1264}, contains questions based on French Wikipedia documents.

We used fine-tuning (on FQuAD) for CamemBERT and the T5 models with 2 epochs, and the \textit{Low-Rank Adaptation} method (LoRA) \cite{hu2021lora} on the LLAMA2 model. For GPT-3.5 and Mixtral 8x7B, respectively, a one- and two-shot prompting approach was used, which involved specifying to the model the requirement for an extraction of the original document with one example of input/output in the expected format.

\subsection{Evaluation}

We evaluate these models on the evaluation corpus with two kinds of metrics: automatic and human metrics.
For the automatic metrics we use the \textit{ROUGE-L} score from the ROUGE toolkit\footnote{We use the google research implementation available \href{https://github.com/google-research/google-research/tree/master/rouge}{here}, with the stemmer and camembert-base tokenizer.} \cite{lin-2004-rouge}. This is a similarity score between the extractive reference answer and the systems output.
For the human metrics we perform a manual annotation of all the systems' output. Annotators were presented with triplets consisting of a context, a question, and an answer. They were  tasked to label each answer as '\textit{correct}', '\textit{partially correct}', or '\textit{incorrect}'. The output from all systems, along with the ground-truth answers, was used to create a total of 14,280 triplets (1,785 triplets per system, including 7 systems and the ground-truth). After removing duplicates in the answers, we obtained a set of 5857 unique triplets, which were then divided into 10 folds and evaluated by 10 human annotators\footnote{All these human annotations as well as systems' output and complexity factors annotations are publicly available : \url{https://anonymous.4open.science/r/CALOR-QA-EMNLP}. Any annotator labels that contradicted the ground-truth labels were reviewed to either correct the reference annotations or adjust the annotators' decisions.}
Two metrics were derived from this manual annotation:
\begin{itemize}
    \item \textit{Hscore}: This metric assigns a score of 1 to answers labeled as \textit{correct}, 0.5 to those labeled as \textit{partially correct}, and 0 to those labeled as \textit{incorrect}.
    \item \textit{Hcorrect}: This metric represents the proportion of answers labeled as \textit{correct} by the annotators for a given system.
\end{itemize}

\begin{table}[ht]
\resizebox{\columnwidth}{!}{%
\begin{tabular}{|l|c|c|c|c|c|}
\hline
\rowcolor{lightgray} Model & adapt & \#param & Rouge-L & Hscore & \% Hcorrect \\ \hline
\textit{CamemBERT} & FT & 335M & 0.82 & 0.85 & 78.9 \\ \hline
\textit{T5-L} & FT & 738M &  0.81 & 0.84 &  78.0 \\ \hline
\textit{FLAN-T5-L} & FT & 783M & 0.80 & 0.85 &  79.2\\ \hline 
\textit{MT5-L}  & FT & 1.2B & 0.80 & 0.84 & 77.5 \\ \hline 
\textit{LLAMA-2}  & LoRA & 7B & 0.69 & 0.78 & 72.2 \\ \hline
\textit{Mixtral-8x7b}  & prompt & 47B & 0.80 & 0.87 & \textbf{82.6} \\ \hline
\textit{GPT 3.5} & prompt & 175B & 0.72 & \textbf{0.88} & 82.5 \\ \hline 
\rowcolor{yellow} \textit{\textbf{ROVER}} & - & - & \textbf{0.84} & \textbf{0.88} & 82.3 \\ \hline
\end{tabular}
}
\caption{Description of the 7 models used in our experiments with their performance in terms of Rouge-L, Hscore and Hcorrect scores. The last line indicates the performance of systems' combination through the ROVER method. \label{tab:rouge-global}}
\end{table}

Overall, the results achieved by the various models are notably lower compared to those showcased on leaderboards of analogous tasks such as SQuAD~\footnote{\url{https://rajpurkar.github.io/SQuAD-explorer}} or MultiRC in SuperGLUE~\footnote{\url{https://super.gluebenchmark.com/leaderboard}}. This discrepancy can be attributed in part to the characteristics of the evaluation corpus and its differences with the adaptation corpus FQuAD as well as the absence of systematic model optimization through hyperparameterization.

The Rouge-L score of the T5-based generation models and the CamemBERT-based classification model are closely aligned, whereas those of the two LLMs, LLAMA-2 and GPT3.5, significantly lag behind.
This comes from the fact that the references in the evaluation corpus are extractive (comprising segments of the original text) and that RougeL inherently leans towards models that merely replicate segments without introducing additional words.
When considering human evaluation, the results are inverted: generative LLMs that are lightly adapted with prompting, that tend to introduce additional elements for presentation or explanation, are preferred by humans and outperform other models on both \textit{Hscore} and \textit{Hcorrect} metrics.

This analysis underscores the necessity for evaluation metrics beyond string similarity between a single reference and the output of a generative model for abstractive tasks. Notably, unlike GPT-3.5 and Mixtral, the LLAMA-2 model's performance remains low in human evaluations. This discrepancy can be attributed to the ineffective LoRA adaptation, despite being monitored using the Rouge-L score.
Although the final Rouge-L score was low, it was comparable to that of GPT-3.5, leading us to initially attribute the low score to the model's abstractive capabilities. However, human evaluation revealed this was not the case. Due to the high cost of human annotation, it was not feasible to use this metric to refine and optimize our adaptation process. Consequently, we exclude the results obtained with LLAMA-2 from now on and use our human metrics instead of Rouge-L.

\subsection{Complexity factors}
We apply the methodology described in section~\ref{sec:method} for partitioning QA examples by complexity and assessing the relevance of the complexity factors describes in section~\ref{sec:complexity-factors}.

\begin{figure}[]
 \begin{center}
  \includegraphics[width=0.5\textwidth]{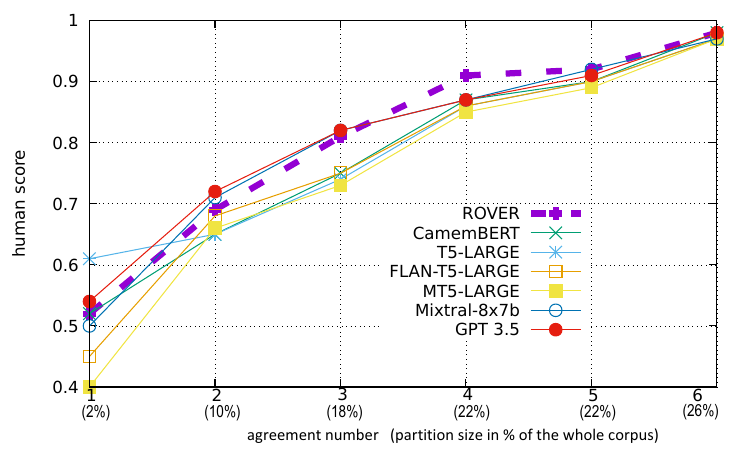} 
  \caption{Performance in Hscore according to the agreement number with the ROVER systems' combination method}
 \label{fig:rover-agree}
 \end{center}
 \end{figure}
 
To sort QA examples by complexity, we utilize the agreement between models, which is assessed using the ROVER score as detailed above. Given that we are working with both extractive and abstractive models, we calculate the \textit{agreement} between the outputs of two models, $M_1(x)$ and $M_2(x)$, for a given input $x$ using the Levenshtein distance, denoted as $dist_L(.)$, between the two strings. The agreement is defined as:
\[
\textit{agree}(M_1,M_2,x) \Leftrightarrow dist_L(M_1(x), M_2(x)) < \alpha
\]
In our experiments, we arbitrarily fixed $\alpha = 5$ to allow strings that differ only by the deletion or addition of a specifier to be considered as agreeing.

The ROVER performance is displayed in the last row of Table \ref{tab:rouge-global}. It performs best for Rouge-L and Hscore metrics and closely approaches the best for Hcorrect.
ROVER forms the basis of our proposed method for sorting QA examples by complexity. By using 6 models in the voting process, we categorize examples into 6 partitions ($P1$ to $P6$) based on the level of agreement among systems. $P1$ contains examples where the 6 systems' outputs differ, while $P6$ includes those where all systems agree.
In Figure \ref{fig:rover-agree}, we plot the Hscore of ROVER and all other models across these 6 partitions. The alignment between the number of agreements and complexity measurement is consistent across all models, with ROVER scores closely mirroring Hscore, which increases nearly linearly with agreement count.
From this curve, we deduce that our evaluation corpus is relatively easy. Nearly half of the corpus ($48\%$, combining $P5$ and $P6$) has an Hscore over 90\% for all models. Of the remaining corpus, 40\% ($P3$ and $P4$) are of moderate complexity, where larger models outperform smaller ones. The final 12\% are the most difficult examples for all models, regardless of their size.

\paragraph{Is complexity linked to semantic relations?} The ROVER partitioning produced reliable clusters but did not clarify why some clusters are more challenging than others. To investigate this, we explore the correlation between semantic relationships linking questions, answers and model performance. Semantic relationships are represented by the frames used to generate the questions (detailed in Section \ref{sec:corpus}). We segmented our corpus into 54 sub-corpora based on the frames, allowing us to evaluate each model's performance for each specific frame.

\begin{figure*}
\begin{center}
 \includegraphics[width=0.9\textwidth]{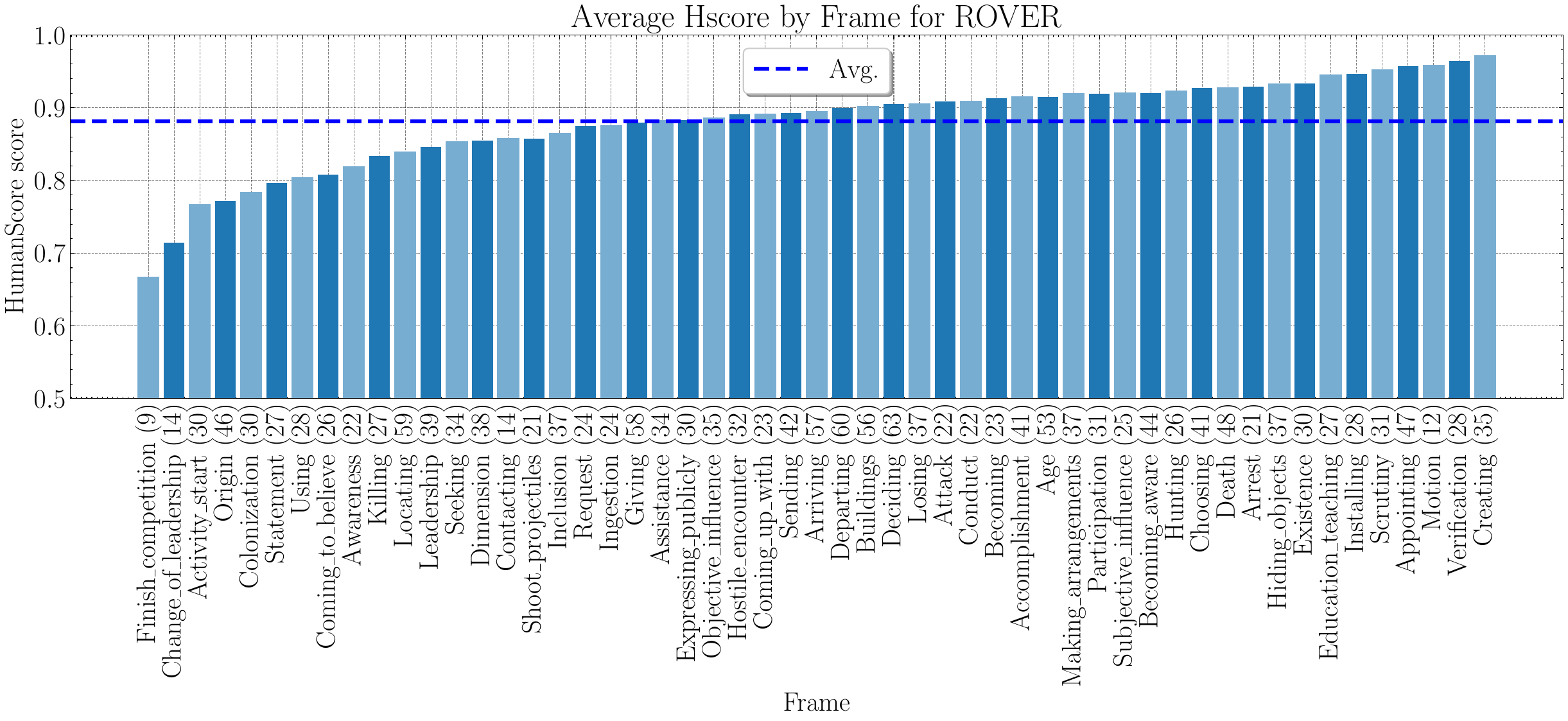} 
 \caption{Performance of ROVER according to each frame sorted by Hscore measure. The number of occurrences of each frame in the corpus is given between brackets}
\label{fig:score-frame}
\end{center}
\end{figure*}

Figure \ref{fig:score-frame} illustrates the distribution of ROVER scores across each frame sub-corpus\footnote{Similar distributions were observed across all models, even if there is some variation in the frame ranking. The figures for all models are in Appendix~\ref{appendix:rouge_h_frame_all}}.
This distribution is non-uniform, validating our intuition that model performance varies with underlying semantic relations.
This brings us to the second step of our method, which involves validating the complexity factors proposed in Section~\ref{sec:complexity-factors}.

\begin{table}[h]
\resizebox{\columnwidth}{!}{%
\begin{tabular}{|c|ccccccc|}
\hline
 &
  \multicolumn{7}{c|}{Complexity factor} \\ \hline
models/factors &
  \multicolumn{1}{c|}{bias} &
  \multicolumn{1}{c|}{coref} &
  \multicolumn{1}{c|}{trigger} &
  \multicolumn{1}{c|}{LU in q} &
  \multicolumn{1}{c|}{dist} &
  \multicolumn{1}{c|}{nb FEs} &
  entropy \\ \hline
size of Ef (\%) &
  \multicolumn{1}{c|}{42\%} &
  \multicolumn{1}{c|}{6\%} &
  \multicolumn{1}{c|}{37 \%} &
  \multicolumn{1}{c|}{45\%} &
  \multicolumn{1}{c|}{30\%} &
  \multicolumn{1}{c|}{59\%} &
  50\% \\ \hline
CamemBERT &
  \multicolumn{1}{c|}{-1} &
  \multicolumn{1}{c|}{\textbf{-7}} &
  \multicolumn{1}{c|}{-1} &
  \multicolumn{1}{c|}{-2} &
  \multicolumn{1}{c|}{-1} &
  \multicolumn{1}{c|}{\textbf{-4}} &
  -1 \\ \hline
T5 &
  \multicolumn{1}{c|}{0} &
  \multicolumn{1}{c|}{\textbf{-9}} &
  \multicolumn{1}{c|}{-1} &
  \multicolumn{1}{c|}{-2} &
  \multicolumn{1}{c|}{-3} &
  \multicolumn{1}{c|}{\textbf{-4}} &
  \textbf{-5} \\ \hline
FLAN &
  \multicolumn{1}{c|}{-1} &
  \multicolumn{1}{c|}{-6} &
  \multicolumn{1}{c|}{-1} &
  \multicolumn{1}{c|}{\textbf{-3}} &
  \multicolumn{1}{c|}{\textbf{-4}} &
  \multicolumn{1}{c|}{-3} &
  \textbf{-5} \\ \hline
MT5 &
  \multicolumn{1}{c|}{-1} &
  \multicolumn{1}{c|}{\textbf{-15}} &
  \multicolumn{1}{c|}{0} &
  \multicolumn{1}{c|}{\textbf{-2}} &
  \multicolumn{1}{c|}{\textbf{-3}} &
  \multicolumn{1}{c|}{\textbf{-4}} &
  \textbf{-4} \\ \hline
GPT-3.5 &
  \multicolumn{1}{c|}{0} &
  \multicolumn{1}{c|}{-2} &
  \multicolumn{1}{c|}{0} &
  \multicolumn{1}{c|}{1} &
  \multicolumn{1}{c|}{-1} &
  \multicolumn{1}{c|}{-1} &
  -2 \\ \hline
mixtral-8x7b &
  \multicolumn{1}{c|}{0} &
  \multicolumn{1}{c|}{-2} &
  \multicolumn{1}{c|}{-2} &
  \multicolumn{1}{c|}{-2} &
  \multicolumn{1}{c|}{-2} &
  \multicolumn{1}{c|}{\textbf{-4}} &
  -1 \\ \hline
ROVER &
  \multicolumn{1}{c|}{1} &
  \multicolumn{1}{c|}{\textbf{-7}} &
  \multicolumn{1}{c|}{0} &
  \multicolumn{1}{c|}{-2} &
  \multicolumn{1}{c|}{-1} &
  \multicolumn{1}{c|}{\textbf{-2}} &
  -2 \\ \hline
\end{tabular}%
}

\caption{Validation results for complexity factors across models, showing $\delta$ values in each cell with statistically significant differences in bold. '\textit{Size}' indicates proportions of partitions $E_{f}$ relative to the total corpus.}
\label{tab:complex}
\end{table}

\paragraph{Evaluation of complexity factors.} 
Table \ref{tab:complex} shows the results for these 7 complexity factors. In each cell, for a model $m$ and a factor $f$, the value corresponds to the impact of $f$ on $m$ expressed by the difference in terms of Hscore $\delta$ presented in Section~\ref{sec:method}.
Values in bold correspond to factors that have validated the Mann-Whitney U test for statistical significance with a 5\% risk. 
This methodology allows us to systematically analyze and quantify the impact of different complexity factors on model performance, providing rigorous statistical validation of observed differences in Hscore between linguistically easier and more complex subgroups.
As we can see, the generic factor $f_{bias}$ corresponding to the link between the frequency of a frame in the adaptation corpus and in the evaluation corpus has very little influence on the results.

Factor $f_{coref}$ shows that resolving co-reference chains is a complexity factor for all models but significantly impacts only smaller models like T5 and MT5. While LLMs also experience some performance loss, it is less significant, indicating their better handling of co-references.

The nature of the Frame trigger ($f_{trigger}$) is a complexity factor for all models but differences are not statistically significant. Factor $f_{LU\;in\;q}$  is validated for all models except GPT-3.5, but significant only for FLAN and MT5. Factor $f_{dist}$ mainly affects smaller models, supporting the idea that LLMs better encode syntactic structures.

Interestingly, the most reliable factors are those intrinsically linked to the semantic relations representing the frames ($f_{nb\;FEs}$ and  $f_{entropy}$) rather than their contextual use. Thus, these two factors can be associated with the measure of semantic ambiguity in question/answer relations.

For example, the \textit{Request} frame has over 20 triggers in the Berkeley Framenet lexicon\footnote{\url{https://framenet.icsi.berkeley.edu/frameIndex}}. In our evaluation corpus, it has 33 occurrences with 6 different triggers, resulting in high entropy and Hscore scores from 0.55 to 0.84 depending on the model.

In contrast, the \textit{Installing} frame, defined as "\textit{An Agent places a Component in a Fixed Location so that the Component is attached and interconnected and thereby functional}" has only two triggers (\textit{install} and \textit{installation}). It has 30 occurrences in our corpus with 2 triggers, low entropy, and Hscore scores from 0.79 to 0.90.

Factor $f_{nb\;FEs}$ shows frames with a low number of Frame Elements in their examples ($\leq 2$). For instance, the \textit{Origin} frame has two 'core' FEs (\textit{Origin} and \textit{Entity}), while the \textit{Giving} and \textit{Contacting} frames have more 'core' and non-core FEs. This aligns with factor $f_{entropy}$, where the \textit{Origin} frame scores below average, while \textit{Giving} is an 'easy' frame.

\begin{figure*}[h!]
 \begin{center}
  \includegraphics[width=1\textwidth]{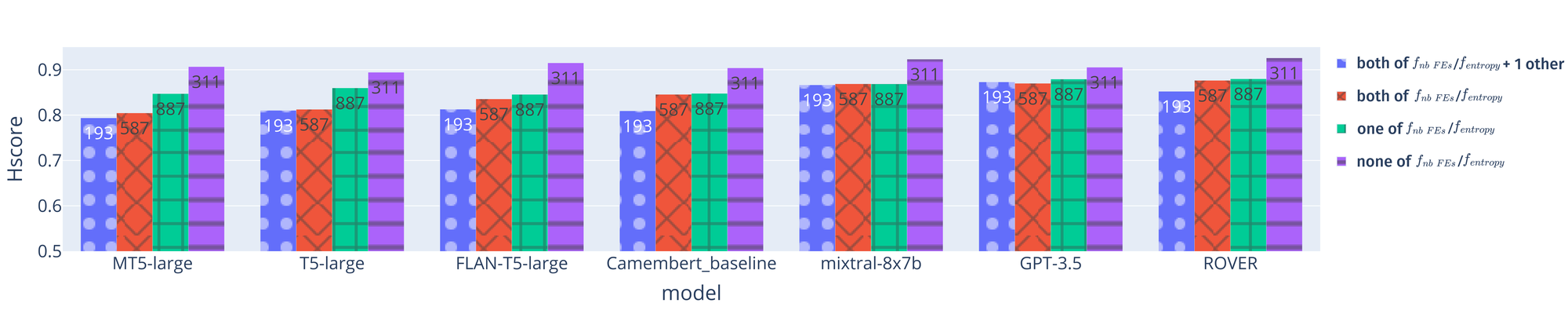} 
  \caption{Hscore on 4 partitions of the evaluation corpus according to combinations of complexity factors}
 \label{fig:score-diff-facile}
 \end{center}
 \end{figure*}

\paragraph{Selecting semantically complex QA examples.} Complexity factors can be used to identify challenging QA examples by considering one or more factors. Our analysis focuses on the most significant factors, $f_{nb\;FEs}$ and $f_{entropy}$. Figure~\ref{fig:score-diff-facile} shows Hscore values for subsets of the corpus categorized by examples influenced by neither, one, or both of these factors, plus any additional factors. Most models exhibit the greatest score disparity between subsets with no factors and those with at least one of $f_{nb\;FEs}$ or $f_{entropy}$. The score difference is minimal between subsets with one factor and those with both, except for T5, MT5, and LLaMA-2.

\begin{table}[h]
\resizebox{\columnwidth}{!}{%
\begin{tabular}{|c|c|c|c|c|c|c|}
\hline
f/P & $P6$ & $P5$ & $P4$ & $P3$ & $P2$ & $P1$ \\ \hline
$P\left(f_{nb\;FEs}\right)$ & 0.52 & 0.56 & 0.62 & 0.64 & 0.62 & 0.80 \\ \hline 
$P\left(f_{entropy}\right)$ & 0.51 & 0.57 & 0.60 & 0.59 & 0.58 & 0.54 \\ \hline
\end{tabular}%
}
\caption{Probability of having the $f_{nb\;FEs}$ and $f_{entropy}$ factors according to the agreement partitions of increasing complexity $P6$ to $P1$}
\label{tab:proba-factor-agreement}
\end{table}

The last step of our analysis is to study if our semantic factors can explain the differences in complexity among the different partitions $P1$ to $P6$ obtained through the ROVER method. Table~\ref{tab:proba-factor-agreement} shows the probabilities of the QA examples in each partition $P$ to have factor $f_{nb\;FEs}$  or $f_{entropy}$.
As can be observed, probabilities for $f_{nb\;FEs}$  and $f_{entropy}$ increase clearly from $P6$ to $P5$ and to a lesser extent from $P5$ to $P4$, indicating that examples with higher semantic ambiguities are more likely to be occurring in the difficult partitions within $P3$ to $P1$.

\section{Experiments with NaturalQA}
\label{sec:natural_qa}

To evaluate the transferability of our two main complexity factors ($f_{nb\;FES}$ and $f_{entropy}$) to other QA datasets and languages, we used a subset of NaturalQA \cite{kwiatkowski2019natural} and the predictions of 48 models provided by HELM \cite{liang2023holistic} on their \textit{natural\_qa\_openbook\_longans}\footnote{\url{https://crfm.stanford.edu/helm/lite/latest/\#/groups/natural_qa_openbook_longans}} scenario. This subset consists of 1,000 examples from the NaturalQA evaluation distribution, each comprising a question, a "short" answer, and the context, which in this case is the corresponding "long" answer from NaturalQA (typically equivalent to a paragraph). For brevity, we present the results of 8 of the 48 models in Table~\ref{tab:complex-naturalQA}\footnote{See result for all model in \ref{appendix:NaturalQA-results}}, selected to represent the full range of mean F1 scores across all models. Additionally, we display the ROVER score estimated across all (48) models.

\paragraph{Applying $f_{entropy}$ to NaturalQA.} For this dataset, lacking an automatic Frame analysis, we used a proxy method: we compiled all potential triggers from Berkeley FrameNet frames and checked their exact presence in the questions. Each question provided a list of triggers and their corresponding frames. Using a custom prompt\footnote{Appendix~\ref{appendix:NaturalQA}}, we employed GPT-3.5 to determine the most appropriate pair for each question.

For example with the question : \textit{How long did the democrats control the house and senate?} we can extract the following list of 11 triggers and their potentials frames : {[('Duration\_description', 'long'), ('Buildings', 'house'), ('Desiring', 'long'), ('Dimension', 'long'), ('Firefighting', 'control'), ('Controller\_object', 'control'), ('Measurable\_attributes', 'long'), ('Containing', 'house'), ('Experimentation', 'control'), ('Being\_in\_control', 'control'), ('Control', 'control')]}. The chosen pair in this case being : ('Being\_in\_control', 'control').\\

We decided to use a proxy via ChatGPT rather than automatic analysis in a semantic framework for several reasons. First, this approach offers simplicity in implementation and scalability to other languages, requiring only hypothesis extraction via keyword search and API calls. Second, our analysis is focused on questions, not paragraphs of text, unlike the typical training data for most semantic frame models, and we had reservations about the models' performance in this context. \\

We performed a manual evaluation on 50 sentences, where two annotators assessed ChatGPT's frame predictions as fully correct, partially correct, or erroneous. The results, shown in Table \ref{tab:human-eval-naturalQA}, demonstrate overall good performances, with some errors observed.

\begin{table}[h]
\centering
\resizebox{\columnwidth}{!}{%
\begin{tabular}{|c|c|c|c|}
\hline
\textbf{Evaluation} & \textbf{Full} & \textbf{Partial} & \textbf{Erroneous} \\ \hline
Frame Prediction    & 57                      & 18                          & 25                    \\ \hline
Frame Elements & 66                          & 22                             & 12                       \\ \hline
\end{tabular}}
\caption{Manual evaluation (in \%) of ChatGPT's frame predictions across 50 random sentences}
\label{tab:human-eval-naturalQA}
\end{table}

Out of 1000 examples, 18 had no extractable triggers and were discarded. On the generated frames by ChatGPT, 35 were unknown from our Frame dictionary and were also discarded. We then assessed our $f_{entropy}$ factor by calculating the median entropy across all Berkeley FrameNet frames.

We computed the variation in F1-score between examples that validated $f_{entropy}$ (high entropy) and those that did not. Across all models, the average difference in performance between these subsets was -3.17 (± 1.82) F1 points, indicating that $f_{entropy}$ is also a significant complexity factor for the NaturalQA benchmark. In Table~\ref{tab:complex-naturalQA}, we present the F1 variation for the 8 selected models as well as for ROVER, showing that most models have a significant F1-lost when considering $f_{entropy}$ examples.

 \begin{table}[]
\centering
\resizebox{\columnwidth}{!}{%
\begin{tabular}{|cc|cc|}
\hline
\multicolumn{2}{|c|}{}                                  & \multicolumn{2}{c|}{Factors}            \\ \hline
\multicolumn{1}{|c|}{models/factors}             & F1   & \multicolumn{1}{c|}{nb FEs}   & entropy          \\ \hline
\multicolumn{1}{|c|}{size of ${E}_{f}$ (\%)}            &      & \multicolumn{1}{c|}{78\%} & 52\%        \\ \hline
\multicolumn{1}{|c|}{google\_text-bison@001}     & 0.81 & \multicolumn{1}{c|}{0}    & -1          \\ \hline
\multicolumn{1}{|c|}{openai\_text-davinci-003}   & 0.77 & \multicolumn{1}{c|}{-4}   & \textbf{-5} \\ \hline
\multicolumn{1}{|c|}{ROVER}  & 0.77 & \multicolumn{1}{c|}{-4}   &    -3       \\ \hline
\multicolumn{1}{|c|}{meta\_llama-3-70b}          & 0.74 & \multicolumn{1}{c|}{-3}   & \textbf{-4} \\ \hline
\multicolumn{1}{|c|}{mistralai\_mixtral-8x7b-32kseqlen} & 0.70 & \multicolumn{1}{c|}{\textbf{-6}} & \textbf{-4} \\ \hline
\multicolumn{1}{|c|}{openai\_gpt-3.5-turbo-0613} & 0.68 & \multicolumn{1}{c|}{-4}   & \textbf{-6} \\ \hline
\multicolumn{1}{|c|}{google\_gemma-7b}           & 0.66 & \multicolumn{1}{c|}{-4}   & -3          \\ \hline
\multicolumn{1}{|c|}{AlephAlpha\_luminous-extended}     & 0.61 & \multicolumn{1}{c|}{\textbf{-7}} & \textbf{-5} \\ \hline
\multicolumn{1}{|c|}{databricks\_dbrx-instruct}  & 0.55 & \multicolumn{1}{c|}{-2}   & 0           \\ \hline
\hline
\rowcolor{lightgray} \multicolumn{1}{|c|}{ROVER}  & 0.77 & \multicolumn{1}{c|}{-4}   &    -3       \\ \hline
\end{tabular}%
}
\caption{Validation results for $f_{nb\;FEs}$ and $f_{entropy}$ across models on naturalQA.
'\textit{Size}' indicates proportions of partitions $E_{f}$ relative to the total corpus.}
\label{tab:complex-naturalQA}
\end{table}

 \paragraph{Applying $f_{nb\;FEs}$ to NaturalQA.} For $f_{nb\;FEs}$, following the prompt method used for $f_{entropy}$, we automatically extracted the FEs based on the previously predicted frames using an empirically developed prompt\footnote{Appendix~\ref{appendix:NaturalQA}}.
 We extracted Frame Elements for 937 examples where frames were successfully predicted out of 961 attempts (24 were discarded due to output issues).

We then categorized these examples based on our $f_{nb\;FEs}$ factor: those with more than 2 FEs are considered easier, while those with 2 or fewer are considered more challenging. Typically, examples with more than 2 FEs score above average, while those with 2 or fewer score below. However, on average, this difference is smaller compared to $f_{entropy}$. Across all models, the average difference in performance between these subsets is -3.84 (± 2.44) F1 points. This may be due to NaturalQA questions being simpler and containing fewer Frame Elements compared to our original corpus, increasing the proportion of challenging examples from 60\% to 78\%.

\section{Related Work}
\label{sec:related}
Our work situates itself within the domain of model evaluation. Our approach contrasts with broad-scale evaluations that span multiple tasks, corpora, and languages \cite{laskar-etal-2023-systematic, liang2023holistic, srivastava2022beyond, brown2020language, wang2019superglue}. It relates to focused studies addressing specific linguistic phenomena such as negations \cite{truong-etal-2022-another, truong-etal-2023-language, zhang-etal-2023-beyond, ravichander-etal-2022-condaqa}, ambiguity in inference tasks \cite{liu-etal-2023-afraid}, and open information extraction \cite{lechelle-etal-2019-wire57}, that utilizes small, meticulously curated datasets to precisely evaluate the capabilities of models for the task. Our study echoes the latter, exploring focused linguistic evaluations.

This study aligns with other efforts evaluating 'closed' LLMs like ChatGPT, accessible only through APIs, on benchmarks such as knowledge-based question answering (KBQA) \cite{tan2023can}. These studies highlight ChatGPT's robust performance across diverse NLP tasks \cite{kocon2023chatgpt, laskar-etal-2023-systematic}, yet also note its potential to lag behind task-specific models. 

Overall, this study pushes the idea that we need a more precise evaluation framework and can be related to other studies such as \cite{ribeiro-etal-2020-beyond} that identify \emph{critical failures} in both commercial and state-of-the-art models by proposing a model and task-agnostic testing methodology or \cite{Gehrmann-Cracked-2023} insisting on the fact that to compare models we need more "\emph{careful annotation process [...] to characterize their output quality and distinguish between them}".

\section{Conclusions}
\label{sec:conclusion}

This paper presents a methodology for identifying intrinsic complexity factors in NLP tasks. Our results reveal that some examples consistently produce lower scores due to their inherent linguistic complexity.
Through an empirical study on a QA task, we identified and validated several factors of semantic complexity, with results directly linked to human evaluations of model predictions. We have also validated these factors on another dataset in another language, confirming their robustness. In addition, we have developed corpora of increasing semantic complexity, suggesting that taking these complexities into account requires more than simply improving the model's parameters.

\section{Limitations}

The main limitation of our study is to have considered a single task, a limited set of languages (French and English) and corpora (CALOR and NaturalQA).
Our focus in this article revolves around the viability of conducting focused, cost-effective studies, requiring less than 100 GPU hours (inclusive of hyperparameter search) and approximately \$10 for the GPT-3.5 API. These studies prioritize linguistic analysis to draw conclusions that extend beyond the specific corpus, task, and language. We believe that such complementary studies have a place in academic Natural Language Processing conferences.

\section*{Acknowledgements}
This project was provided with computer and storage resources by GENCI at IDRIS thanks to the grant 2023-AD011012688R2 on the supercomputer Jean Zay's V100/A100 partition. We would like to thank the reviewers for their comment and feedbacks which has helped us improve the first version of this article to its current state.  

\bibliography{calor}

\section{Appendix}

\appendix

\section{Technical information about the training process and the data}

FQuAD dataset download link : \url{https://fquad.illuin.tech/}

\subsection{Training of CamemBert}
CamemBert was finetuned using the default parameters of the HuggingFace trainer for 4 epochs, with model check-pointing keeping the best overall checkpoint.

\paragraph{\textbf{Training hardware :}}
\begin{verbatim}
GPU : 1 x Tesla V100-SXM2-32GB
\end{verbatim}

\subsection{Training of T5, MT5 and FLAN-T5}
The training was performed using a modified version of this training script script from HuggingFace :  \url{https://github.com/huggingface/transformers/blob/main/examples/pytorch/question-answering/trainer_seq2seq_qa.py}

\textbf{Training parameters }are bellow, all other parameters are the \textbf{default} one of the HuggingFace trainer (\textbf{transformers} installation from source at commit \hyperlink{https://github.com/huggingface/transformers/commit/686c68f64c9d0181bd54d4d2e2446543c3eca1fa}{686c68f64c9d0181bd54d4d2e2446543c3eca1fa}). 

\begin{verbatim}
{
    "max_seq_length": 512,
    "adafactor": true,
    "learning_rate" : 3e-05,
    "num_train_epochs" : 2,
    "evaluation_strategy": "steps",
    "metric_for_best_model": "f1",
    "load_best_model_at_end": true,
    "seed": 260,
    "max_answer_length": 40
}

\end{verbatim}

\paragraph{\textbf{Data format  :}}
\begin{verbatim}
"question: {question} 
contexte : {context}" 
\end{verbatim}

\paragraph{\textbf{Training hardware :}}
\begin{verbatim}
GPU : 1 x Tesla V100-SXM2-32GB
\end{verbatim}

\paragraph{\textbf{Training time :}}
\begin{itemize}
\item T5 $\approx$ 2h15mn
\item MT5 $\approx$ 2h30mn
\item FLAN-T5 $\approx$ 2h30mn
\end{itemize}

In total, a few run of tests ($\approx$ 12) for the prompt, optimizer and learning rate were done with similar running times.

The \textbf{inference} time vary a bit between model and is $\approx$ 30mn.

\subsection{Adaptation of llama2-7b}

The LoRA adaptation was performed using \url{https://github.com/huggingface/peft} library, with the config given bellow.

\begin{verbatim}
LoraConfig(
    r=32,
    lora_alpha=64,
    target_modules=["q_proj", "v_proj"],
    lora_dropout=0.1,
    bias="none",
    task_type="CAUSAL_LM",
)
\end{verbatim}

The modified training argument are given bellow, the rest are default.

\begin{verbatim}
transformers.TrainingArguments(
    per_device_train_batch_size=1,
    gradient_accumulation_steps=4,
    num_train_epochs=1,
    learning_rate=2e-4,
    fp16=True,
    save_total_limit=3,
    logging_steps=1,
    max_steps=80,
    optim="paged_adamw_32bit",
    lr_scheduler_type="cosine",
    warmup_ratio=0.05,
)
\end{verbatim}

\paragraph{\textbf{Prompt format  :}}
The prompt was constructed with the same three examples randomly selected from FQuAD for both training and inference.
\begin{verbatim}
Below is a paragraph of text, paired with a 
question. Extract the sequence of words in 
the article that answers the following 
question, or answer NULL if there are no 
answers.


### Paragraph:
Après le tournage, Hal B. Wallis [...]
### Question:
Qui ne peut pas se libérer pour la scène 
envisagée par Wallis ?
### Answer:
"Claude Rains"


### Paragraph:
Riquet étudie de façon approfondie  [...]
### Question:
Quel est l'un des points sur lequel le projet 
de Riquet reste imprécis ?
Answer:
"tracé du canal"


### Paragraph:
Dans cet intervalle de 31 jours, [...]
### Question:
Combien sont-ils à être frappés ?
Answer:
"quelques-uns"
\end{verbatim}

\paragraph{\textbf{Training hardware :}}
\begin{verbatim}
GPU : 1 x GPU Nvidia A100-80GB
\end{verbatim}

\paragraph{\textbf{Training and inference time :}}
\begin{itemize}
\item training $\approx$ 70sec
\item inference $\approx$ 17mn
\end{itemize}

\subsection{rouge-L results and significativity for the complexity factors}

\begin{table}[h!]
\resizebox{\columnwidth}{!}{%
\begin{tabular}{c|ccccccc|}
\cline{2-8}
 &
  \multicolumn{7}{c|}{Complexity factor} \\ \hline
\multicolumn{1}{|c|}{models/factors} &
  \multicolumn{1}{c|}{bias} &
  \multicolumn{1}{c|}{coref} &
  \multicolumn{1}{c|}{trigger} &
  \multicolumn{1}{c|}{LU in q} &
  \multicolumn{1}{c|}{dist} &
  \multicolumn{1}{c|}{nb FEs} &
  entropy \\ \hline
\multicolumn{1}{|c|}{size of Ef (\%)} &
  \multicolumn{1}{c|}{42\%} &
  \multicolumn{1}{c|}{6\%} &
  \multicolumn{1}{c|}{37\%} &
  \multicolumn{1}{c|}{45\%} &
  \multicolumn{1}{c|}{12\%} &
  \multicolumn{1}{c|}{59\%} &
  46\% \\ \hline
\multicolumn{1}{|c|}{CamemBERT} &
  \multicolumn{1}{c|}{\cellcolor[HTML]{FFFFFF}{\color[HTML]{000000} -1}} &
  \multicolumn{1}{c|}{\cellcolor[HTML]{FFFFFF}{\color[HTML]{000000} -4}} &
  \multicolumn{1}{c|}{\cellcolor[HTML]{FFFFFF}{\color[HTML]{000000} -1}} &
  \multicolumn{1}{c|}{\cellcolor[HTML]{FFFFFF}{\color[HTML]{000000} \textbf{-2}}} &
  \multicolumn{1}{c|}{\cellcolor[HTML]{FFFFFF}{\color[HTML]{000000} \textbf{-7}}} &
  \multicolumn{1}{c|}{\cellcolor[HTML]{FFFFFF}{\color[HTML]{000000} \textbf{-3}}} &
  \cellcolor[HTML]{FFFFFF}{\color[HTML]{000000} -1} \\ \hline
\multicolumn{1}{|c|}{T5} &
  \multicolumn{1}{c|}{\cellcolor[HTML]{FFFFFF}{\color[HTML]{000000} -1}} &
  \multicolumn{1}{c|}{\cellcolor[HTML]{FFFFFF}{\color[HTML]{000000} -9}} &
  \multicolumn{1}{c|}{\cellcolor[HTML]{FFFFFF}{\color[HTML]{000000} -2}} &
  \multicolumn{1}{c|}{\cellcolor[HTML]{FFFFFF}{\color[HTML]{000000} -1}} &
  \multicolumn{1}{c|}{\cellcolor[HTML]{FFFFFF}{\color[HTML]{000000} \textbf{-7}}} &
  \multicolumn{1}{c|}{\cellcolor[HTML]{FFFFFF}{\color[HTML]{000000} \textbf{-5}}} &
  \cellcolor[HTML]{FFFFFF}{\color[HTML]{000000} \textbf{-2}} \\ \hline
\multicolumn{1}{|c|}{FLAN} &
  \multicolumn{1}{c|}{\cellcolor[HTML]{FFFFFF}{\color[HTML]{000000} -2}} &
  \multicolumn{1}{c|}{\cellcolor[HTML]{FFFFFF}{\color[HTML]{000000} -4}} &
  \multicolumn{1}{c|}{\cellcolor[HTML]{FFFFFF}{\color[HTML]{000000} -3}} &
  \multicolumn{1}{c|}{\cellcolor[HTML]{FFFFFF}{\color[HTML]{000000} -2}} &
  \multicolumn{1}{c|}{\cellcolor[HTML]{FFFFFF}{\color[HTML]{000000} -4}} &
  \multicolumn{1}{c|}{\cellcolor[HTML]{FFFFFF}{\color[HTML]{000000} \textbf{-5}}} &
  \cellcolor[HTML]{FFFFFF}{\color[HTML]{000000} \textbf{-3}} \\ \hline
\multicolumn{1}{|c|}{MT5} &
  \multicolumn{1}{c|}{\cellcolor[HTML]{FFFFFF}{\color[HTML]{000000} 0}} &
  \multicolumn{1}{c|}{\cellcolor[HTML]{FFFFFF}{\color[HTML]{000000} -13}} &
  \multicolumn{1}{c|}{\cellcolor[HTML]{FFFFFF}{\color[HTML]{000000} -1}} &
  \multicolumn{1}{c|}{\cellcolor[HTML]{FFFFFF}{\color[HTML]{000000} -1}} &
  \multicolumn{1}{c|}{\cellcolor[HTML]{FFFFFF}{\color[HTML]{000000} \textbf{-10}}} &
  \multicolumn{1}{c|}{\cellcolor[HTML]{FFFFFF}{\color[HTML]{000000} \textbf{-4}}} &
  \cellcolor[HTML]{FFFFFF}{\color[HTML]{000000} -2} \\ \hline
\multicolumn{1}{|c|}{llama-2} &
  \multicolumn{1}{c|}{\cellcolor[HTML]{FFFFFF}{\color[HTML]{000000} 0}} &
  \multicolumn{1}{c|}{\cellcolor[HTML]{FFFFFF}{\color[HTML]{000000} -3}} &
  \multicolumn{1}{c|}{\cellcolor[HTML]{FFFFFF}{\color[HTML]{000000} -1}} &
  \multicolumn{1}{c|}{\cellcolor[HTML]{FFFFFF}{\color[HTML]{000000} 3}} &
  \multicolumn{1}{c|}{\cellcolor[HTML]{FFFFFF}{\color[HTML]{000000} -3}} &
  \multicolumn{1}{c|}{\cellcolor[HTML]{FFFFFF}{\color[HTML]{000000} \textbf{-7}}} &
  \cellcolor[HTML]{FFFFFF}\textbf{-2} \\ \hline
\multicolumn{1}{|c|}{GPT-3.5} &
  \multicolumn{1}{c|}{\cellcolor[HTML]{FFFFFF}{\color[HTML]{000000} 0}} &
  \multicolumn{1}{c|}{\cellcolor[HTML]{FFFFFF}{\color[HTML]{000000} \textbf{4}}} &
  \multicolumn{1}{c|}{\cellcolor[HTML]{FFFFFF}{\color[HTML]{000000} -1}} &
  \multicolumn{1}{c|}{\cellcolor[HTML]{FFFFFF}{\color[HTML]{000000} 0}} &
  \multicolumn{1}{c|}{\cellcolor[HTML]{FFFFFF}{\color[HTML]{000000} -4}} &
  \multicolumn{1}{c|}{\cellcolor[HTML]{FFFFFF}{\color[HTML]{000000} -4}} &
  \cellcolor[HTML]{FFFFFF}{\color[HTML]{000000} \textbf{-3}} \\ \hline
\multicolumn{1}{|c|}{mixtral-8x7b} &
  \multicolumn{1}{c|}{0} &
  \multicolumn{1}{c|}{1} &
  \multicolumn{1}{c|}{-2} &
  \multicolumn{1}{c|}{-1} &
  \multicolumn{1}{c|}{\cellcolor[HTML]{FFFFFF}-5} &
  \multicolumn{1}{c|}{\cellcolor[HTML]{FFFFFF}\textbf{-6}} &
  \cellcolor[HTML]{FFFFFF}0 \\ \hline
\end{tabular}%
}
\caption{Complexity factor validation results with the Rouge-L score. Each box contains the $\delta$ value of each factor for each model. Bold indicate statistically significant differences. The \textit{size} line displays the proportions of the $E_{f}$ partitions relative to the total size of the corpus.}
\label{tab:complex-ROUGE-L}
\end{table}

\subsection{Extraction of Frames and Frame Element on NaturalQA}
\label{appendix:NaturalQA}
\paragraph{\textbf{Prompt for Frame extraction :}}
\begin{verbatim}
From a list of (frame, lexical unit) from 
FrameNet, predict which is the most likely 
for the given question. Only answer with 
the correct (frame, lexical unit) pair.
List : {list}
Question : {question}
\end{verbatim}

\paragraph{\textbf{Prompt for Frame Element extraction :}}
\begin{verbatim}
From a FrameNet (frame , lu/trigger) pair 
and a context extract the corresponding 
Frame Elements from the given question. 
The LU can't be a FE.  Output a json.
Pair : {pair}
Question : {question}
\end{verbatim}

\subsection{Complexity factor examples}
\label{appendix:factor_examples}
\paragraph{\mbox{Number of Arguments in the Frame} ($f5$) : }

\textbf{Easy (more FEs in context, here > 2) :}\\
Comment est mort Kleitarchos en 341 ?\\
(\textit{How did Kleitarchos die in 341?})\\
Quand les congrès de Zimmerwald et de Kiental ont-ils commencé le processus de renversement de l'ordre établi ?\\
(\textit{When did the Zimmerwald and Kiental congresses begin the process of overthrowing the established order?})\\
Lors de la bataille d'Actium, Caius Sosius a dirigé quelle partie de la flotte ?\\
(\textit{At the battle of Actium, which part of the fleet did Caius Sosius command?})\\
En quelle année Silvestras Žukauskas a-t-il été étudiant à l'école des cadets d'infanterie de Wilna~?\\
(\textit{In what year was Silvestras Žukauskas a student at the Wilna Infantry Cadet School?})\\

\textbf{Hard (less FEs in context, here 2) :}\\
Qu'est-ce qui est caché ?\\
(\textit{What's hidden?})\\
Quand les Russes attaquent-ils ?\\
(\textit{When do the Russians attack?})\\
Quel est le sujet ?\\
(\textit{What's the subject?})\\
Who shoots the ammunition?\\
(\textit{Who shoots the ammunition?})\\
Qui a découvert de nouvelles techniques de création ?\\
(\textit{Who's discovered new creative techniques?})\\

\subsection{Annotator compensation}

The human annotators are volunteer PhD students from the same laboratory (from different teams to the authors). They were paid 45€ via gift vouchers, as our country's legislation does not allow direct pay-per-task remuneration.

\subsection{HumanScore results per frame for all models}
\label{appendix:rouge_h_frame_all}

\begin{figure*}[!h]
\centering
  \includegraphics[width=0.9\textwidth]{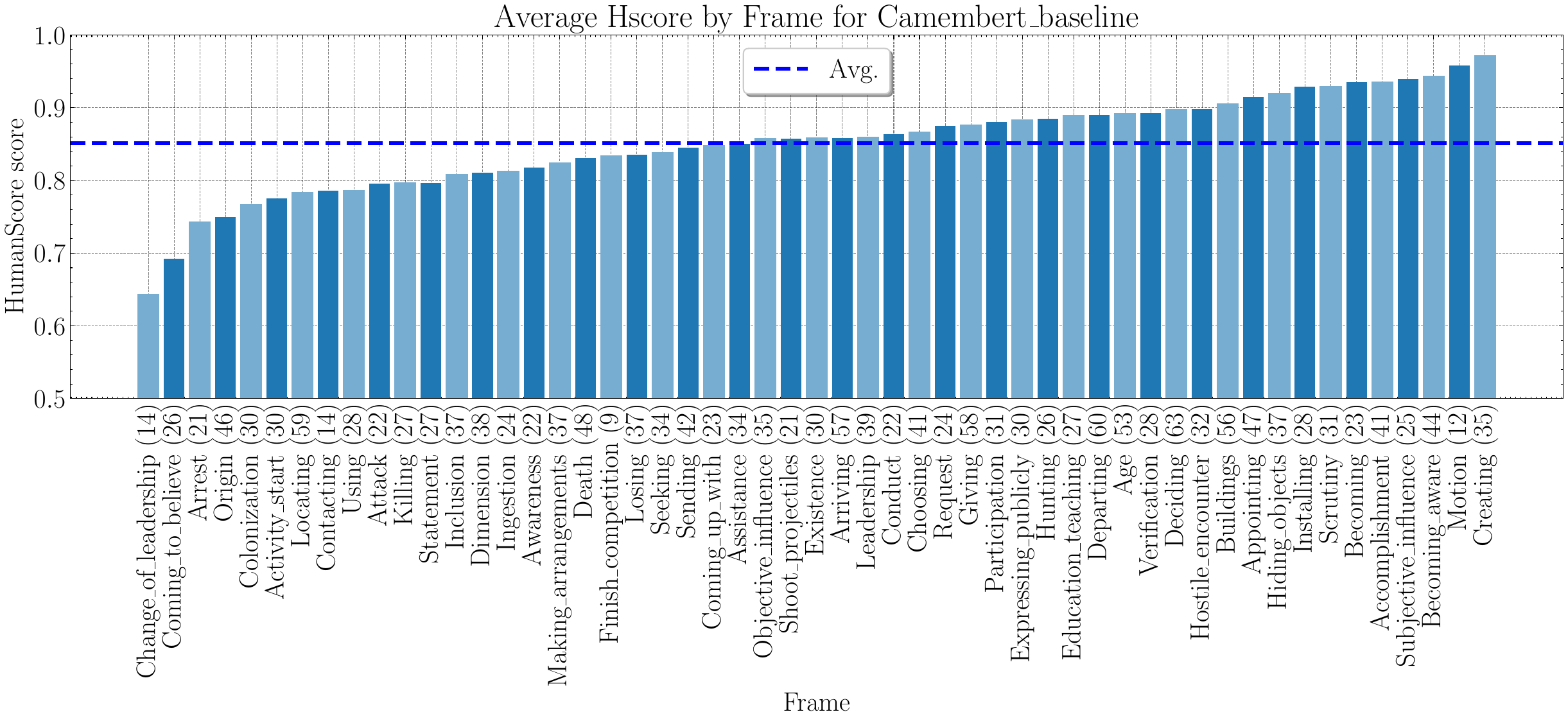} 
 \end{figure*}

\begin{figure*}[!h]
\centering
  \includegraphics[width=0.9\textwidth]{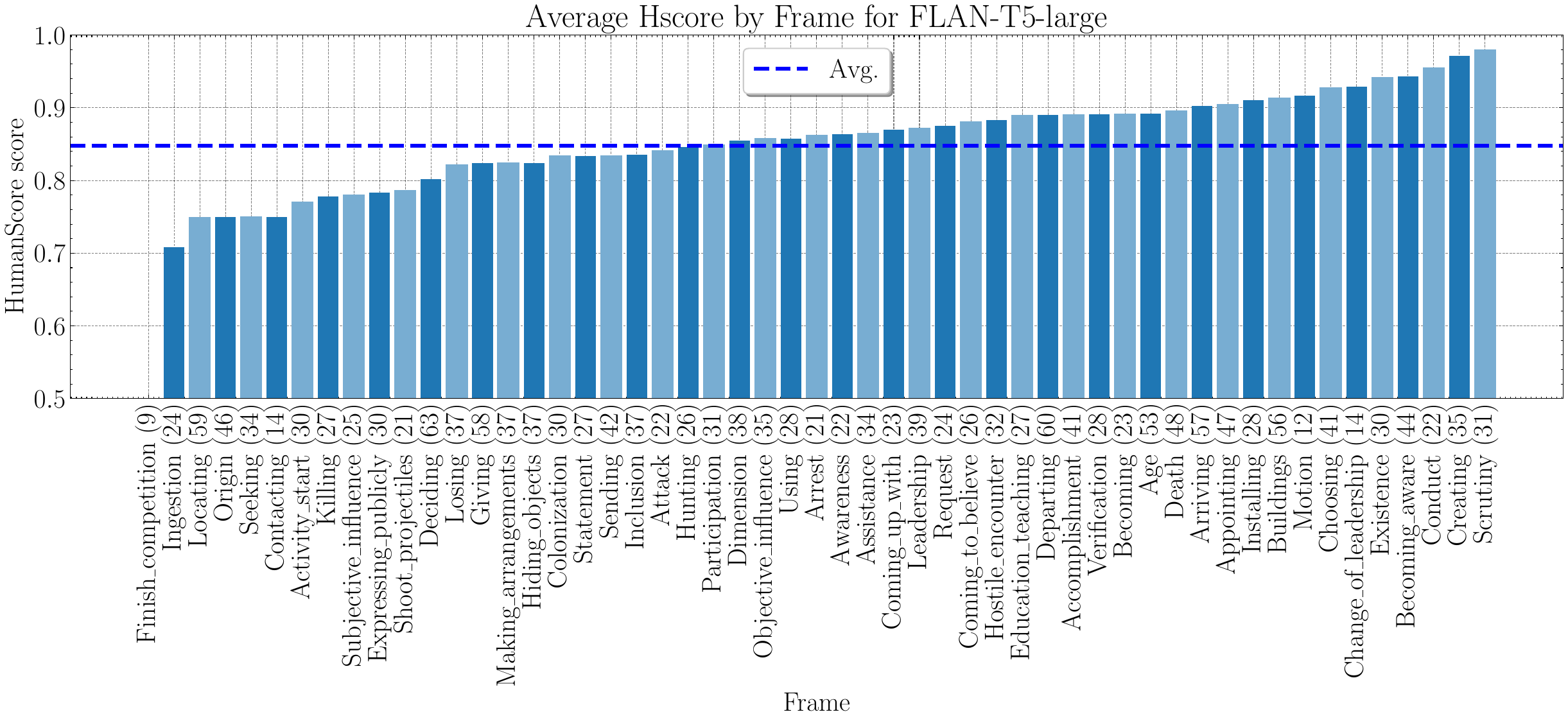} 
 \end{figure*}

\begin{figure*}[!h]
\centering
  \includegraphics[width=0.9\textwidth]{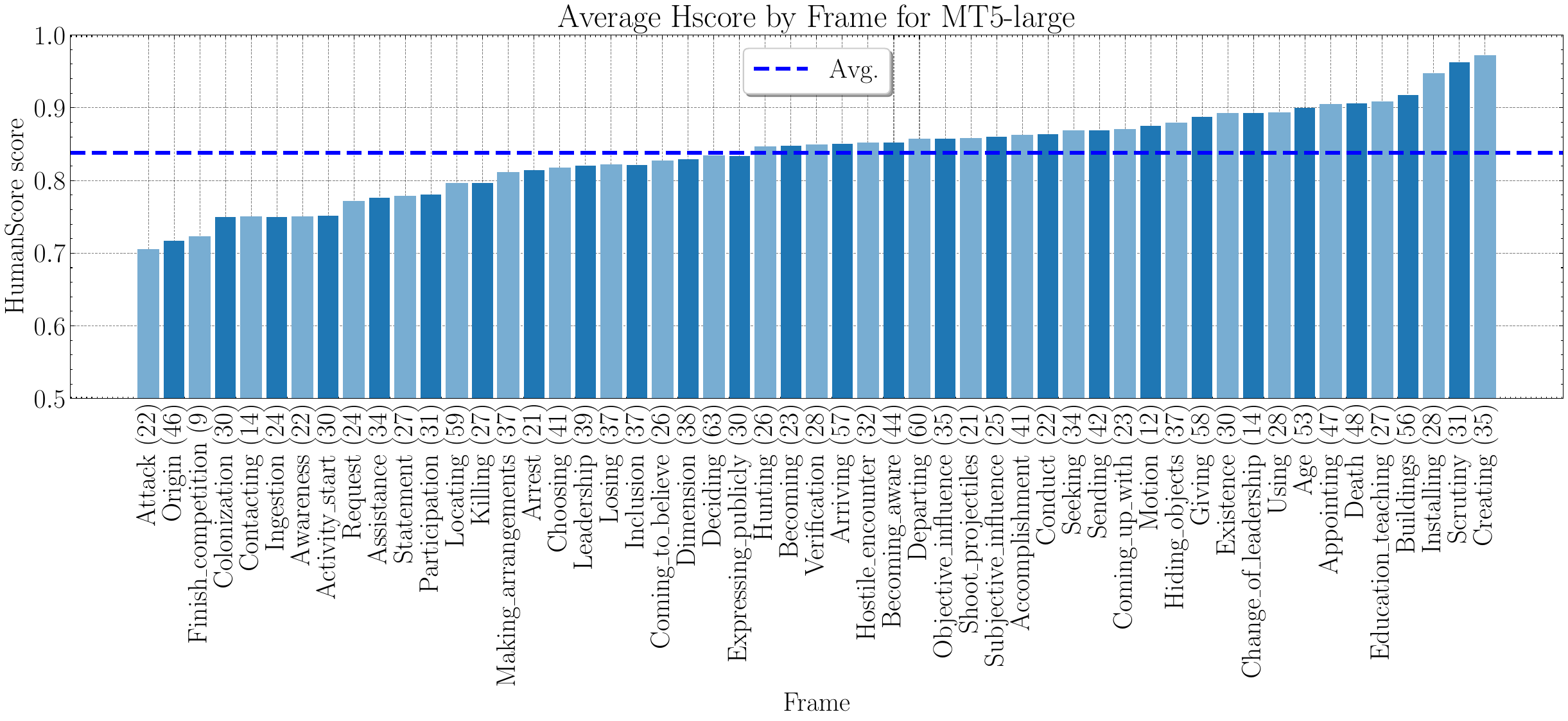} 
 \end{figure*}

\begin{figure*}[!h]
\centering
  \includegraphics[width=0.9\textwidth]{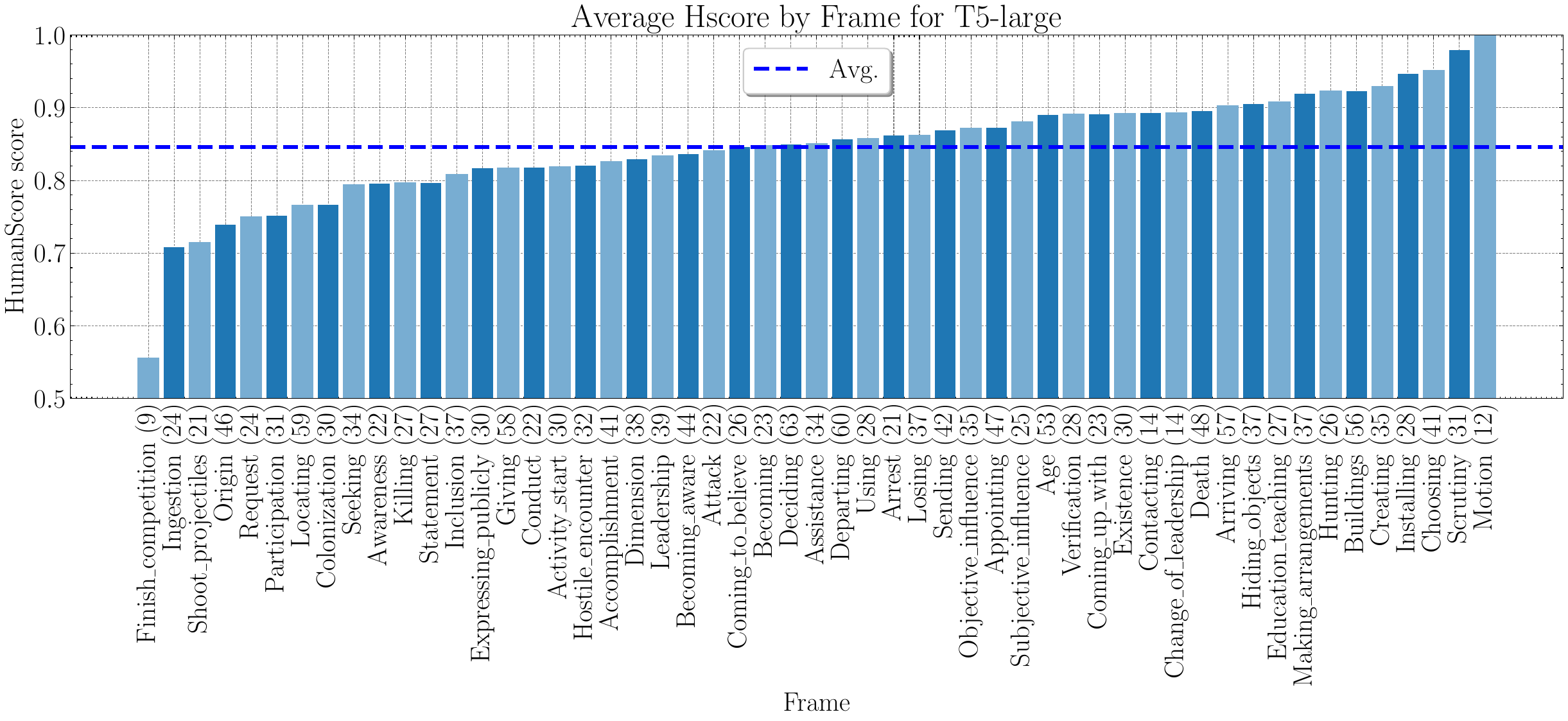} 
 \end{figure*}

\begin{figure*}[!h]
\centering
  \includegraphics[width=0.9\textwidth]{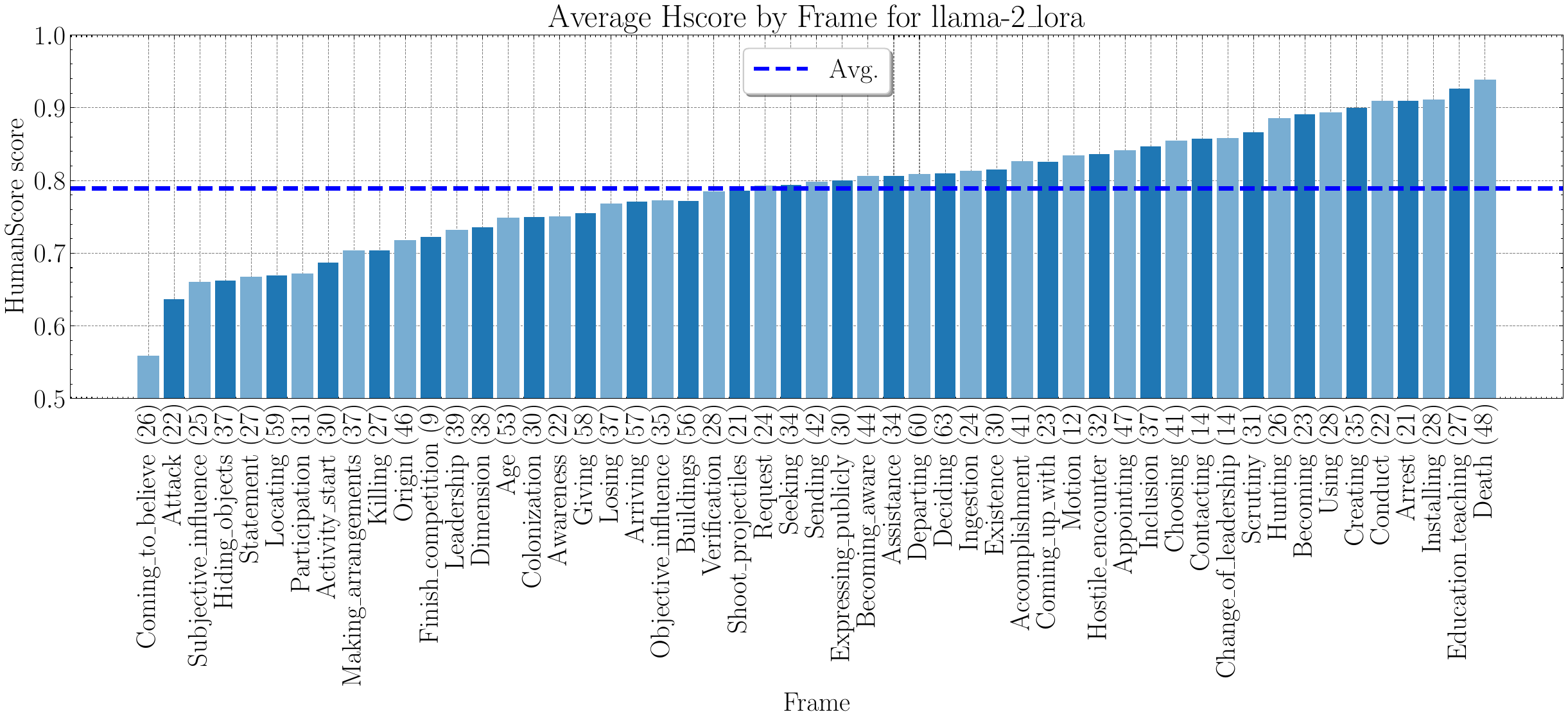} 
 \end{figure*}

 \begin{figure*}[!h]
\centering
  \includegraphics[width=0.9\textwidth]{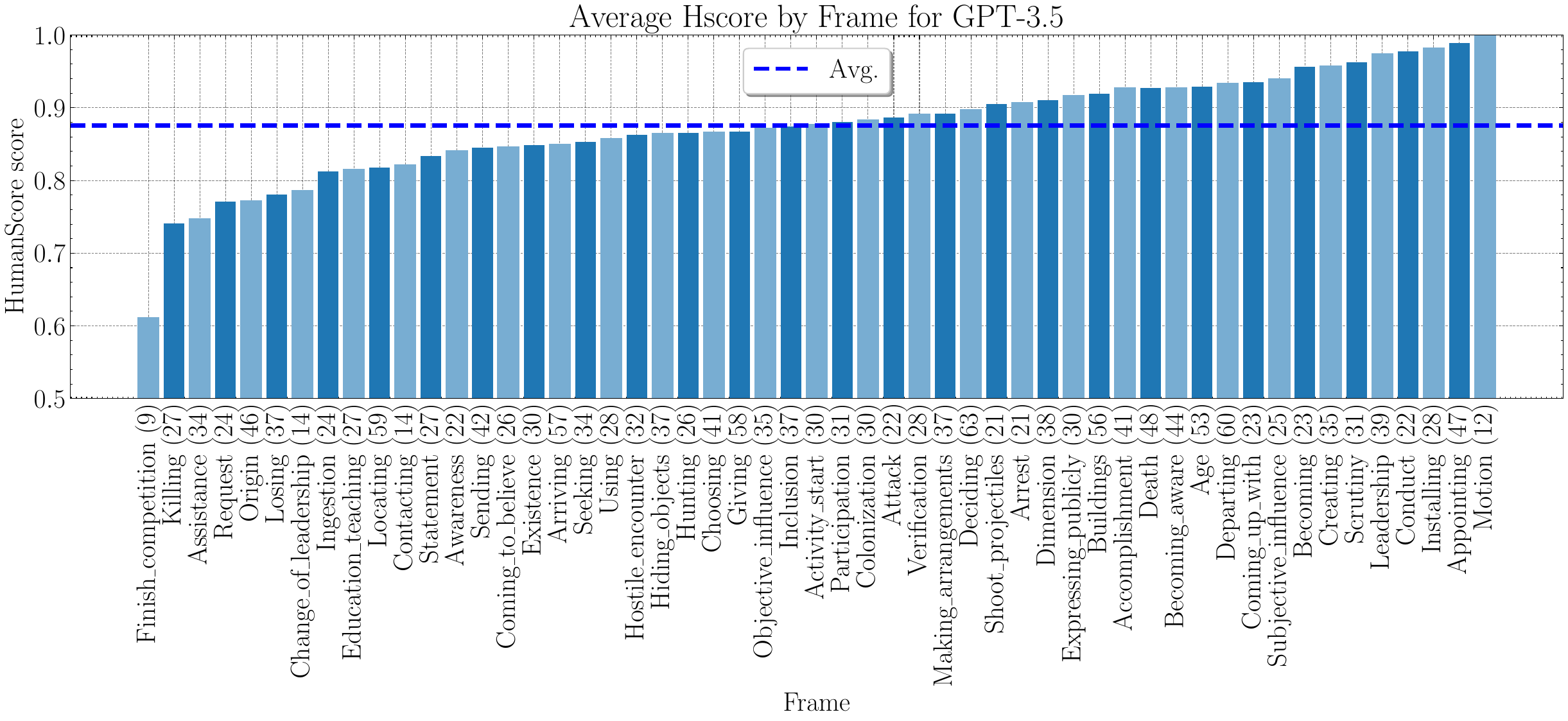} 
 \end{figure*}
 
 \begin{figure*}[!h]
\centering
  \includegraphics[width=0.9\textwidth]{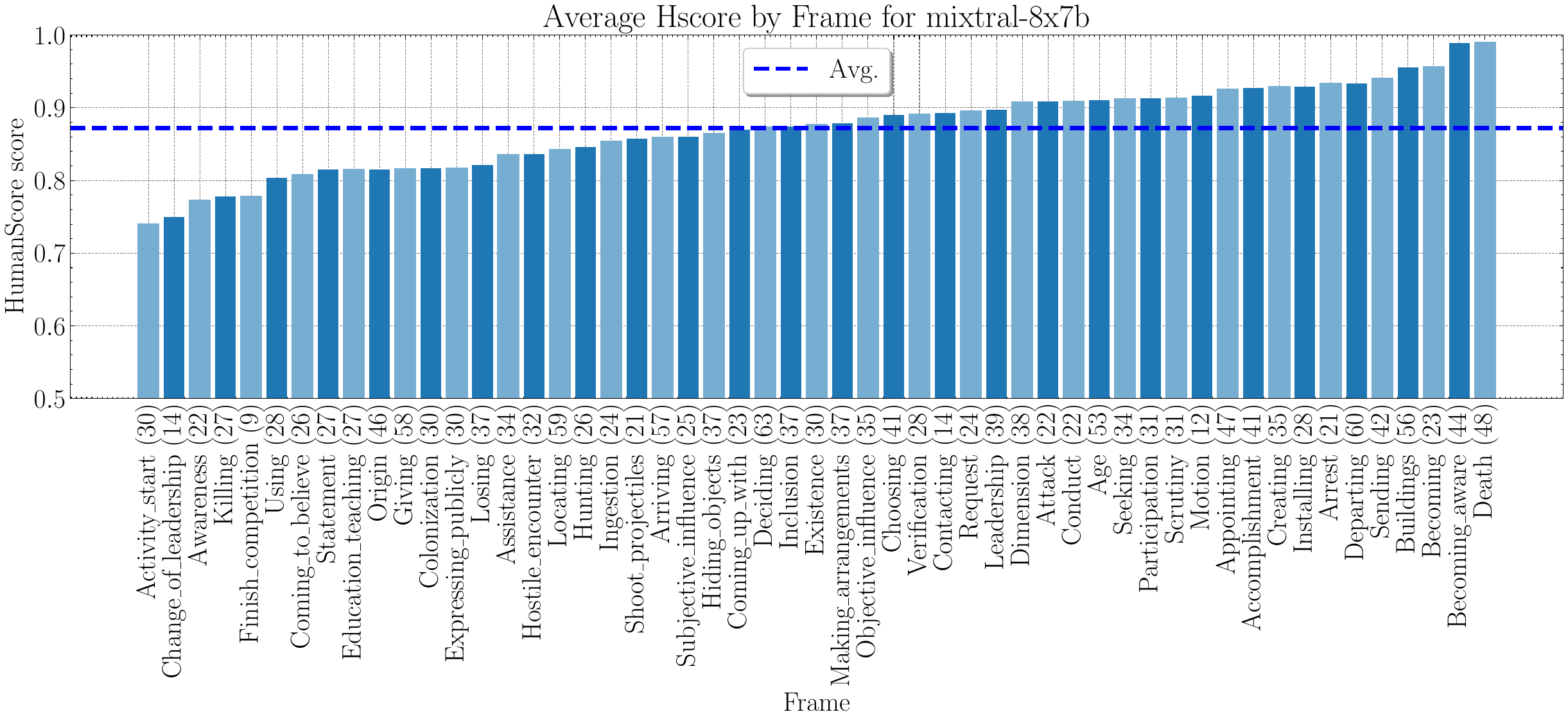} 
 \end{figure*}

\subsection{Result on all model of naturalQA for $f5$ and $f6$}
\label{appendix:NaturalQA-results}

 \begin{figure*}[!h]
\centering
  \includegraphics[width=0.9\textwidth]{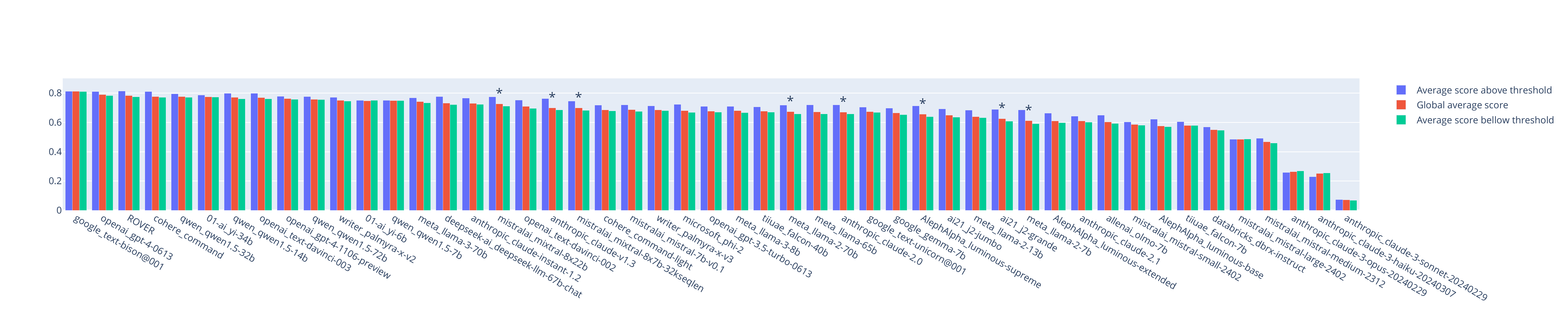} 
  \caption{$f5$ complexity factor on all the examples of naturalQA
  }
 \end{figure*}

 \begin{figure*}[!h]
\centering
  \includegraphics[width=0.9\textwidth]{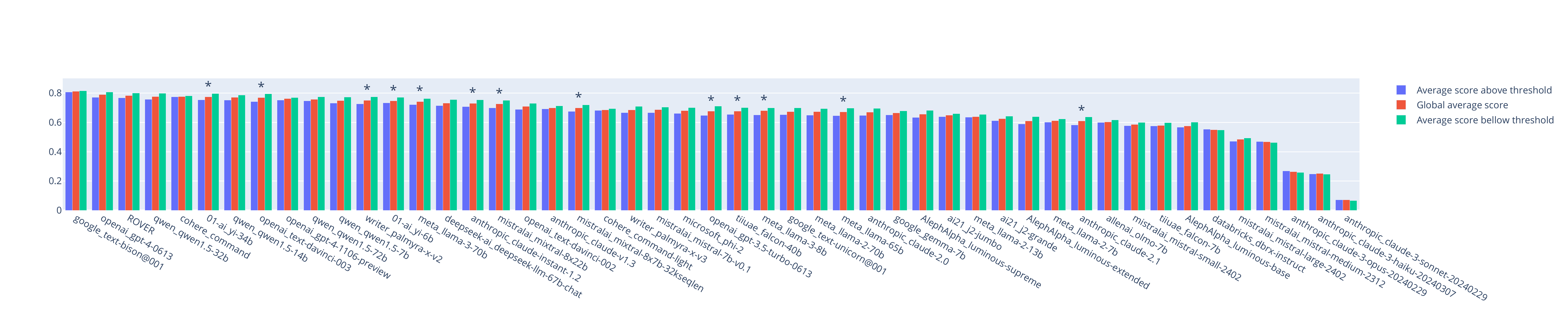} 
  \caption{$f6$ complexity factor on all the examples of naturalQA}
 \end{figure*}

\end{document}